# HISSNET: SOUND EVENT DETECTION AND SPEAKER IDENTIFICATION VIA HIERARCHICAL PROTOTYPICAL NETWORKS FOR LOW-RESOURCE HEADPHONES

*N Shashaank*[1,2], *Berker Banar*[1,3], *Mohammad Rasool Izadi*[1],
*Jeremy Kemmerer*[1], *Shuo Zhang*[1], *Chuan-Che (Jeff) Huang*[1]

[1] Bose Corporation, MA, USA
[2] Department of Computer Science, Columbia University, NY, USA
[3] School of Electronic Engineering and Computer Science, Queen Mary University of London, UK

## ABSTRACT

Modern noise-cancelling headphones have significantly improved users' auditory experiences by removing unwanted background noise, but they can also block out sounds that matter to users. Machine learning (ML) models for sound event detection (SED) and speaker identification (SID) can enable headphones to selectively pass through important sounds; however, implementing these models for a user-centric experience presents several unique challenges. First, most people spend limited time customizing their headphones, so the sound detection should work reasonably well out of the box. Second, the models should be able to learn over time the specific sounds that are important to users based on their implicit and explicit interactions. Finally, such models should have a small memory footprint to run on low-power headphones with limited on-chip memory. In this paper, we propose addressing these challenges using HiSSNet (Hierarchical SED and SID Network). HiSSNet is an SEID (SED and SID) model that uses a hierarchical prototypical network to detect both *general* and *specific* sounds of interest and characterize both alarm-like and speech sounds. We show that HiSSNet outperforms an SEID model trained using non-hierarchical prototypical networks by 6.9 – 8.6%. When compared to state-of-the-art (SOTA) models trained specifically for SED or SID alone, HiSSNet achieves similar or better performance while reducing the memory footprint required to support multiple capabilities on-device.

*Index Terms* — Sound event detection (SED), speaker identification (SID), hierarchical prototypical networks

## 1. INTRODUCTION

The advancement of noise cancellation technology for headphones and earbuds has enabled people to focus on their work and enjoy music even in a noisy environment, whether it is outdoors in a crowded plaza or indoors in a busy house. However, current noise-cancelling headphones can also block out important sounds that matter to users, such as a nearby person trying to get their attention, a passing car or truck while walking on the street, or someone trying to ring the user's doorbell. An ideal solution is to embed sound event detection (SED) and speaker identification (SID) machine learning (ML) models directly in noise-cancelling headphones, enabling it to block out unwanted sounds while passing through sounds that are important to the user.

Neural networks are a common ML model architecture used in SED systems [1] to detect *general* sounds of interest that require user attention, such as speech, appliance sounds, etc. [2], [3]. These models can detect sounds with no user input, which is favorable for consumer products as users typically spend limited time customizing their devices (ex. ~25% of users adjust equalizer settings in the Bose Music app). However, it is currently difficult for neural networks to differentiate between *specific* sounds that are important or unimportant; for example, speech from a spouse or child is important, while speech from a television is not as important.

Few-shot learning models [4]–[6] can be trained to differentiate between specific sounds and personalized to the user's environment based on explicit interactions (e.g., record real-world sound samples) and potentially implicit interactions (e.g., head movement when wearing a headphone) [7]. However, such models require significant user inputs to start functioning, so it has been used in accessibility applications for deaf and hard of hearing users, as they are more willing to spend time recording audio samples to create a personalized SED system [4]. While it is possible to train separate machine learning models to detect both general sounds that may matter to users and prioritize specific sounds as a system learns user preferences, the memory required to run multiple models concurrently makes it more challenging to deploy on low-power embedded devices that have minimal (1 – 4 MB) on-chip storage [8].

In this paper, we present three contributions. First, we propose HiSSNet, a hierarchical prototypical network [6] for general and personalized SEID (SED and SID). With no user input, HiSSNet can store pre-trained embeddings to classify general sounds of interest, and with some real-world samples the model can be personalized to create new embeddings and detect specific sounds from the user's environment. Second, we propose a hierarchical class ontology for HiSSNet to define relationships between general and specific sound classes and separate sounds that require different user attention levels. Finally, we show that HiSSNet can outperform a standard prototypical network (ProtoNet) trained on SEID by 6.9 – 8.6% and can achieve similar or better performance compared to state-of-the-art (SOTA) models that are specifically trained for SED or SID alone while reducing the memory footprint for running SEID on-device.

## 2. METHODS

The model architecture for HiSSNet was adapted from the hierarchical prototypical network [6], a few-shot learning model that can predict classes using a limited set of examples and incorporates a hierarchical structure to define relationships between general and specific classes. In a few-shot classification task, we are given a labeled support set $S$ with $N$ examples. Each support example $x_S \in S$ is labeled with a class $k \in K$ from a set of classes $K$; the subset of $S$ labeled with class $k$ is defined as $S_k$. The objective of the task is to predict a target class $k \in K$ for each example $x_Q \in Q$ in an unlabeled query set $Q$ with $N$ examples. A neural network encoder $f_\theta$ is used to transform the support and query sets into a latent embedding space.

### 2.1. Hierarchical Prototypical Network

A hierarchical prototypical network [6] computes class prototypes using the mean of support example feature embeddings for each level $h = 0 \ldots H$ of a tree $T$ with height $H$. Starting from the lowest level $h = H$, the prototypes $\hat{x}_s$ for level $h$ are aggregated based on the hierarchical relationships defined in $T$ to form meta-prototypes at level $h - 1$:

$$c_{T_k^{h-1}} = \frac{1}{|S_{T_k^h}|} \sum_{\hat{x}_S \in S_{T_k^h}} f_\theta(\hat{x}_S) \quad (1)$$

In our notation, 0 is the top-most level of the hierarchy containing general sound class labels, and $H$ is the bottom-most level of the hierarchy containing specific sound class labels. The classes in $k \in K$ are used to form prototypes for level $H$.

For classifying a query sample $x_Q$, each level of the hierarchy is treated as a separate few-shot classification task. Using a distance function $d$, a softmax is applied to the distances between each query sample feature embedding and class prototype or meta-prototype, and a probability distribution is created for each level:

$$p_\theta(y_Q = T_k^h | x_Q) = \frac{\exp\left(-d\left(f_\theta(x_Q), c_{T_k^h}\right)\right)}{\sum_{c'_{T_k^h}} \exp\left(-d\left(f_\theta(x_Q), c'_{T_k^h}\right)\right)} \quad (2)$$

The class with the closest prototype distance to the query sample is assigned as the predicted label. The learnable parameters in the neural network encoder $\theta$ are optimized during training with stochastic gradient descent.

A hierarchical prototypical network is optimized using a weighted loss function $L$ that trains the model to incorporate the relationships between general and specific sound classes into the latent embedding space. The loss function is an exponentially decaying sum of cross-entropy loss terms for each level of the hierarchy:

$$L = \sum_{h=0}^{H} \sum_{x_Q \in Q} e^{\alpha \cdot h} \log p_\theta(y_Q = T_k^h | x_Q) \quad (3)$$

A hyperparameter $\alpha$ is used to adjust the weight of the optimization objective between general (upper-level) and specific (lower-level) classes. $\alpha > 0$ gives more weight to specific classes, $\alpha = 0$ equally weights general and specific classes, and $\alpha < 0$ gives more weight to general classes.

### 2.2. HiSSNet

To perform general and personalized SEID, we used the hierarchical prototypical network for HiSSNet, and for the neural network encoder we selected the MobileNetV2 architecture [9], a convolutional neural network that efficiently runs on the latest-generation of ultra-low-power microcontrollers [8] and has previously been used in a few-shot learning model for SED [4]. Additionally, we tested two different distance functions for $d$ in HiSSNet: squared Euclidean distance [10] and angular distance [11], which combines a cosine similarity metric with learnable scale and bias parameters. We ultimately selected the Euclidean distance function as it achieved better accuracy in our testing (results not shown).

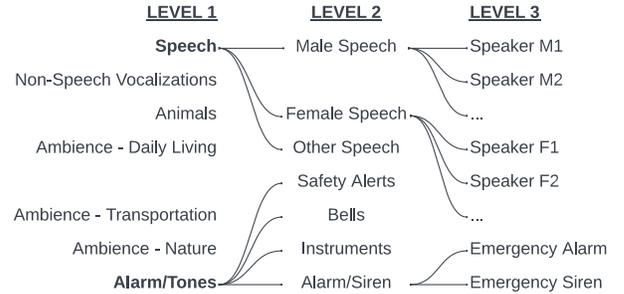

**Fig. 1:** Partial definition of the taxonomy used for training HiSSNet. Bolded classes are sounds important to users.

We created a custom, three-level sound class hierarchy (Fig. 1) for training HiSSNet and evaluating all other models. The hierarchy was created with an emphasis on speech and alarm-like sounds (including alarms, appliance timers, door bells, phone notifications, etc.), as prior studies [2], [3] have shown that users are primarily interested in these two categories of sound classes. To create a taxonomy of classes to use for the hierarchy, we filtered specific sound classes in our selected datasets to match the hierarchy and manually aggregated them into individual labels. In total, the hierarchy contains 7 top-level classes, 19 middle-level classes, and 1319 lower-level classes, with 56 classes for SED and 1263 classes for SID. We have made both the sound class hierarchy and the corresponding taxonomy available online at https://github.com/Bose/HiSSNet-Hierarchy.

The inclusion of a concrete sound class hierarchy in HiSSNet will enable the model to function with varying levels of input when deployed to low-resource headphones. In the base case scenario, pre-trained embeddings for upper-level classes (e.g. Speech, Alarm) can be stored on-device to allow HiSSNet to classify general sounds of interest with no user input. Over time, the model can be trained to differentiate specific sounds of interest based on implicit user interactions; for example, a nearby speaker (Speaker X) can be annotated with a positive response if the user turns their head in response to the sound or if the headphones detect self-voice activity. Finally, if users choose to record audio samples and provide them for personalization, HiSSNet can improve upon the pre-trained embeddings and provide a better experience based on the user's surroundings.

We trained HiSSNet on all sound classes to ensure the encoder learns the latent embedding space for the full hierarchical ontology without assuming the model will generalize to unseen classes during testing. We believe it is better to treat novel classes as "unknown" via an open set approach so the headphone system can take the corresponding action.

## 3. EXPERIMENTAL SETUP

### 3.1. Datasets

For all experiments, we compiled single sound class recordings for SED and SID from 7 different datasets – ESC50 [12], TUT Sound Events 2016 [13], TAU MREAL [14], FSD50K [15], BBC [16], VCTK [17], and LibriSpeech [18]. For TUT and TAU, we included recordings with single sound classes and excluded recordings with overlapping sound classes. In ESC50, FSD50K, BBC, VCTK, and LibriSpeech, all recordings contain a single sound class by default. In total, our collated SEID dataset contains ~211.6K audio files amounting to ~673.4 hours of audio. The dataset was randomly split into 10 stratified folds; 9 folds were used for K-fold cross validation during training, while 1 fold was used for post-training evaluation. All HiSSNet and baseline models were trained and evaluated on the same fold configuration, and all metrics reported in the Results were computed on the post-training evaluation fold.

Audio files were downsampled to 16 kHz before being downmixed to mono, single-channel waveforms, and 1-second segments were randomly sampled from each file. To ensure that the segments were non-silent and contained sound events, a simple dB threshold was applied during segment sampling. We then applied two separate data augmentation transformations for our models to be robust to variations in background noise and context. The first transformation was to mix the audio segments with background noise from an ambient sound scene, such as indoors in a restaurant or an outdoor park. The SNR between the sample and the ambient sound scene was randomly selected between 10 dB to 20 dB. The second transformation was to apply a reverb unit with randomly generated parameters. After data augmentation, each segment was converted into a 64-bin log-Mel spectrogram with a 32ms window and 10ms hop size, creating 64 × 97 bin spectrograms which were finally passed as input to HiSSNet and baseline models.

### 3.2. Model Training Procedure

HiSSNet and ProtoNet models were trained using episodic batches, where each batch contains a random subset of 12 sound classes from the taxonomy and 5 recordings for each class. To sufficiently balance training on both SED and SID classes, we created three different batch configurations: SED only, SED & SID, and SID only. The configuration for each batch was randomly selected during training, with a weight distribution of 60%/20%/20%, respectively. Each epoch contained 100 episodic batches, and the model was trained for 1000 epochs. We used the Adam optimizer [19] with an L2 regularization penalty for performing stochastic gradient descent in all models.

All HiSSNet models were trained on the full SEID dataset, while SOTA baseline models were trained on SED-specific or SID-specific subsets of the dataset. For the SED baselines, we implemented a dilated convolutional recurrent neural network (CRNN) [20] and a non-hierarchical ProtoNet [4], and for the SID baselines we implemented a non-hierarchical ProtoNet [11]. The SED baselines were trained on the data subset from ESC50, TUT, TAU, FSD50K and BBC, while the SID baselines were trained on the data subset from VCTK and LibriSpeech. The dilated CRNN was trained using standard batch processing with a batch size of 128.

### 3.3. Evaluation Metrics

We used the same episodic batch settings (12 sound classes, 5 recordings per class, 100 episodes) from training for all evaluation experiments except the equal error rate (EER) comparison on SID. Our primary evaluation metric is the per-segment accuracy on each level (L1, L2, L3) in the sound class hierarchy, with an emphasis on the top-most level (L1), which captures the performance with no user configuration, and the bottom-most level (L3), which captures the performance after a user provides 5 audio recordings for 12 sound classes. To further evaluate the models' performance on upper-level sound classes in the hierarchy, we also use a hierarchical mistake (HM) metric [6] that measures the height to the lowest common ancestor between the predicted and ground truth labels of an incorrectly classified sample. For the SID task, we ran a separate experiment to compute the EER by running 100 trials and randomly sampling 1000 speaker pairs for each trial, which was adapted from a previous study [11]. All tables report the mean ± SEM of the 100 episodic batches or EER trials.

## 4. RESULTS

### 4.1. Weighting Specific Sound Classes Improves Accuracy

We conducted three cross-validation ablation experiments with HiSSNet to compare different weighting schemes ($\alpha = \{-1, 0, 1\}$) in the hierarchical loss function and analyze the impact of selecting different cross-validation schemes in our dataset. Our results (Table 1) show that giving more weight to specific sound classes ($\alpha = 1$) in the loss function helps the model achieve the best performance across all levels, which is in agreement with the original hierarchical prototypical network study that focused on instrument classification [6]. HiSSNet achieves 93.5% on general SED (L1), 91.3% on specific SED (L3), and 88.9% on specific SID (L3).

The results in Table 1 also show minimal variance in both the accuracy (< 2.5%) and the hierarchical mistake (< 0.1) across different cross-validation training schemes. While HiSSNet with $\alpha = 0$ and HiSSNet with $\alpha = -1$ perform marginally better on the hierarchical mistake, our post-evaluation Wilcoxon signed-rank and Kruskal-Wallis tests (data not shown) revealed that the differences are not statistically significant.

| Model | Metric | SED | SID |
|---|---|---|---|
| HiSSNet $\alpha = -1$ | L1 Acc. | 89.5 ± 1.0% | - |
| | L2 Acc. | 84.6 ± 1.3% | 95.4 ± 0.5% |
| | L3 Acc. | 77.6 ± 1.3% | 73.4 ± 2.2% |
| | H. Mistake | 2.15 ± 0.05 | 1.13 ± 0.01 |
| HiSSNet $\alpha = 0$ | L1 Acc. | 91.7 ± 1.2% | - |
| | L2 Acc. | 89.6 ± 1.3% | 95.9 ± 0.6% |
| | L3 Acc. | 85.9 ± 1.3% | 84.8 ± 1.6% |
| | H. Mistake | 2.26 ± 0.05 | 1.15 ± 0.02 |
| HiSSNet $\alpha = 1$ | L1 Acc. | **93.5 ± 1.1%** | - |
| | L2 Acc. | **93.1 ± 1.2%** | **96.7 ± 0.3%** |
| | L3 Acc. | **91.3 ± 1.2%** | **88.9 ± 1.3%** |
| | H. Mistake | 2.29 ± 0.09 | 1.15 ± 0.02 |

**Table 1:** Cross-validation ablation experiments of HiSSNet with different hierarchical loss weighting schemes.

### 4.2. Training ProtoNet with Hierarchy Achieves Better Accuracy

We compared HiSSNet with a non-hierarchical ProtoNet trained for joint SEID. A key drawback of the non-hierarchical ProtoNet is that it does not force similar sounds that require different attention levels to be in different embedding regions (e.g., phone notifications vs. plate clinking sound while washing dishes). If we pre-compute prototypes for upper-level general classes using such networks based on sound classes that matter to users, it is possible that other unimportant sounds with similar characteristics can end up being close to these upper-level prototypes. We thus hypothesize that HiSSNet will perform better than the non-hierarchical ProtoNet, especially for the top-level accuracy. Our results in Table 2 confirm this hypothesis, as HiSSNet outperforms SEID ProtoNet by 8.6% on general SED (L1), 6.9% on specific SED (L3), and 8.3% on specific SID (L3).

| Model | Metric | SED | SID |
|---|---|---|---|
| SEID ProtoNet | L1 Acc. | 85.5 ± 5.6% | - |
| | L2 Acc. | 85.1 ± 5.7% | 93.3 ± 4.5% |
| | L3 Acc. | 84.8 ± 5.2% | 80.0 ± 6.2% |
| | H. Mistake | 2.46 ± 0.31 | 1.17 ± 0.14 |
| HiSSNet $\alpha = 1$ | L1 Acc. | **94.1 ± 3.1%** | - |
| | L2 Acc. | **93.2 ± 3.6%** | **96.5 ± 3.3%** |
| | L3 Acc. | **91.7 ± 4.0%** | **88.3 ± 5.0%** |
| | H. Mistake | 2.30 ± 0.52 | 1.16 ± 0.22 |

**Table 2:** Comparison with non-hierarchical ProtoNet.

### 4.3. Sound Event Detection Baseline Comparison

We compared HiSSNet with two SOTA models trained specifically for SED [4], [20]. While HiSSNet enables both general and specific SEID in one model to minimize storage and memory, our initial hypothesis was that it could potentially suffer from lower accuracy compared to models trained separately for individual tasks.

To our surprise, our results (Table 3) show that HiSSNet achieves better accuracy compared to the SOTA models, with an improvement of 5.2% on general SED (L1) and 3.2% on specific SED (L3) compared to SED-only ProtoNet. HiSSNet also achieved an improvement of 14.8% over the dilated CRNN, which was only evaluated for specific SED due to the classification architecture.

| Model | #Params | #MAC | Metric | SED |
|---|---|---|---|---|
| SED-only ProtoNet [4] | 2.2M | 45.3M | L1 Acc. | 88.9 ± 5.2% |
| | | | L2 Acc. | 88.7 ± 4.5% |
| | | | L3 Acc. | 88.5 ± 4.9% |
| | | | H. Mistake | 2.47 ± 0.37 |
| SED-only Dilated CRNN [20] | 0.9M | 87.2M | L1 Acc. | - |
| | | | L2 Acc. | - |
| | | | L3 Acc. | 76.9 ± 6.4% |
| | | | H. Mistake | - |
| HiSSNet $\alpha = 1$ | 2.2M | 45.3M | L1 Acc. | **94.1 ± 3.1%** |
| | | | L2 Acc. | **93.2 ± 3.6%** |
| | | | L3 Acc. | **91.7 ± 4.0%** |
| | | | H. Mistake | 2.30 ± 0.52 |

**Table 3:** Comparison with SED baselines.

### 4.4. Speaker Identification Baseline Comparison

We also compared HiSSNet with a SOTA SID approach [11] that uses ProtoNet. For the encoder, instead of the proposed ThinResNet, we applied the MobileNetV2 architecture like our other ProtoNet models as the number of MAC operations is ~10x smaller. Our results (Table 4) show that HiSSNet achieves close performance, with a drop of 6.7% in specific SID (L3) and 0.88% in EER. This minimal decrease in performance is likely acceptable in our problem context of noise-cancelling headphones due to the smaller negative cost of false positives and false negatives, which will temporarily reduce (false positive) or not affect (false negative) the noise cancellation level.

| Metric | SID-only ProtoNet | HiSSNet $\alpha = 1$ |
|---|---|---|
| L2 Acc. | **97.8 ± 2.3%** | 96.5 ± 3.3% |
| L3 Acc. | **95.0 ± 3.5%** | 88.3 ± 5.0% |
| EER | **0.59 ± 0.03%** | 1.47 ± 0.09% |

**Table 4:** Comparison with an SID baseline.

### 5. CONCLUSION

This paper presents HiSSNet, a hierarchical prototypical network for joint SED and SID. This model can detect sounds using no user input, learn and differentiate the specific sounds that matter to users over time based on implicit and explicit user interaction, and run within the on-chip memory constraints of embedded devices. It allows us to address the unique challenges encountered when embedding sound event detection abilities in low-resource noise-cancelling headphones. We show that HiSSNet outperforms non-hierarchical ProtoNet when trained on an SEID dataset. Furthermore, HiSSNet achieves close or better performance as SOTA models trained solely for SED or SID while reducing the memory required to support multiple capabilities on-device.